\PassOptionsToPackage{dvipsnames}{xcolor}
\documentclass{article}

\usepackage{microtype}
\usepackage{graphicx}
\usepackage{subfig}
\usepackage{booktabs} 
\usepackage{url}
\usepackage[breakable,skins]{tcolorbox}
\usepackage{cancel}
\usepackage[title]{appendix}
\usepackage{layouts}
\usepackage{enumerate}
\usepackage{lipsum}
\usepackage{nicefrac}
\usepackage[group-separator={,},output-decimal-marker={.}]{siunitx}

\usepackage{hyperref}

\definecolor{rsfp}{rgb}{0.4980392156862745, 0.4980392156862745, 0.4980392156862745}
\definecolor{gpfp}{rgb}{0.8901960784313725, 0.4666666666666667, 0.7607843137254902}
\definecolor{lafp}{rgb}{0.5490196078431373, 0.33725490196078434, 0.29411764705882354}
\definecolor{lamf}{rgb}{0.5803921568627451, 0.403921568627451, 0.7411764705882353}
\definecolor{lat5}{rgb}{0.8392156862745098, 0.15294117647058825, 0.1568627450980392}
\definecolor{lagpt2}{rgb}{0.17254901960784313, 0.6274509803921569, 0.17254901960784313}
\definecolor{lallama}{rgb}{1.0, 0.4980392156862745, 0.054901960784313725}
\definecolor{lat5chem}{rgb}{0.12156862745098039, 0.4666666666666667, 0.7058823529411765}
\definecolor{boliftllama}{rgb}{0.7372549019607844, 0.7411764705882353, 0.13333333333333333}
\definecolor{boliftgpt}{rgb}{0.09019607843137255, 0.7450980392156863, 0.8117647058823529}

\def\gpfp{{\color{gpfp}GP}}
\def\lafp{{\color{lafp}LA}}
\def\lamf{{\color{lamf}MolFormer}}
\def\latfive{{\color{lat5}T5}}
\def\lagpt{{\color{lagpt2}GPT-2-M}}
\def\lallama{{\color{lallama}Llama-2-7b}}
\def\latfivechem{{\color{lat5chem}T5-Chem}}
\def\boliftllama{{\color{boliftllama}Llama-2-7b}}
\def\boliftgpt{{\color{boliftgpt}GPT-4}}

\usepackage[accepted]{icml2024}

\usepackage{amsmath}
\usepackage{amssymb}
\usepackage{mathtools}
\usepackage{amsthm}

\newtcolorbox{summarybox}{
detach title, fonttitle=\bfseries, coltitle=WildStrawberry, before upper={\tcbtitle\hspace{0.5em}}, title=Summary:, colback=White, colframe=black, rounded corners, left=0pt, right=0pt, top=0pt, bottom=0pt, boxsep=5pt, boxrule=1pt
}

\usepackage[capitalize,noabbrev]{cleveref}

\usepackage{multirow}

\usepackage{tikz}
\usepackage{tikzscale}

\usetikzlibrary{external}
\usetikzlibrary{cd}
\tikzcdset{
  arrow style=tikz,
  diagrams={>={Straight Barb}}
}

\usetikzlibrary{patterns,positioning,arrows,shapes,shapes.geometric,shadows,fit}
\usetikzlibrary{decorations,decorations.pathmorphing,intersections,backgrounds}

\usepackage{pgfplots}
\DeclareUnicodeCharacter{2212}{−}
\usepgfplotslibrary{groupplots,dateplot}
\usetikzlibrary{shapes.arrows}
\pgfplotsset{compat=newest}
\usepgfplotslibrary{groupplots}

\pgfplotsset{compat=1.11,
  /pgfplots/ybar legend/.style={
  /pgfplots/legend image code/.code={%
    \draw[##1,/tikz/.cd,yshift=-0.25em]
    (0cm,0cm) rectangle (3pt,0.8em);},
  },
}

\definecolor{color0}{rgb}{0.12156862745098,0.466666666666667,0.705882352941177}
\definecolor{color1}{rgb}{0.0901960784313725,0.745098039215686,0.811764705882353}
\definecolor{color2}{rgb}{0.737254901960784,0.741176470588235,0.133333333333333}
\definecolor{color3}{rgb}{0.890196078431372,0.466666666666667,0.76078431372549}
\definecolor{color4}{rgb}{0.549019607843137,0.337254901960784,0.294117647058824}
\definecolor{color5}{rgb}{0.580392156862745,0.403921568627451,0.741176470588235}
\definecolor{color6}{rgb}{0.83921568627451,0.152941176470588,0.156862745098039}
\definecolor{color7}{rgb}{0.172549019607843,0.627450980392157,0.172549019607843}

\usepackage{amsmath,amssymb,amsfonts,bm,amsthm,mathtools}

\makeatletter
\def\th@plain{%
  \thm@notefont{}
  \itshape 
}
\def\th@definition{%
  \thm@notefont{}
  \normalfont 
}
\makeatother

\def\1{\bm{1}}

\newcommand{\Dtest}{\mathcal{D_{\mathrm{cand}}}}

\def\vtheta{{\boldsymbol{\theta}}}

\def\vh{{\bm{h}}}

\def\vw{{\bm{w}}}
\def\vx{{\bm{x}}}

\def\mA{{\bm{A}}}
\def\mB{{\bm{B}}}

\def\mH{{\bm{H}}}

\def\mJ{{\bm{J}}}
\def\mK{{\bm{K}}}

\def\mO{{\bm{O}}}

\def\mQ{{\bm{Q}}}

\def\mV{{\bm{V}}}
\def\mW{{\bm{W}}}
\def\mX{{\bm{X}}}

\def\mSigma{{\boldsymbol{\varSigma}}}

\DeclareMathAlphabet{\mathsfit}{\encodingdefault}{\sfdefault}{m}{sl}
\SetMathAlphabet{\mathsfit}{bold}{\encodingdefault}{\sfdefault}{bx}{n}

\newcommand{\R}{\mathbb{R}}

\DeclareMathOperator*{\argmax}{arg\,max}

\newcommand{\vomg}{\boldsymbol{\omega}}

\newcommand{\vgamma}{\boldsymbol{\gamma}}

\newcommand{\N}{\mathcal{N}}

\renewcommand{\R}{\mathbb{R}}

\newcommand{\D}{\mathcal{D}}

\newcommand{\abs}[1]{\vert #1 \vert}
\newcommand{\inv}{{-1}}

\newcommand{\X}{\mathcal{X}}
\newcommand{\Y}{\mathcal{Y}}
\newcommand{\V}{\mathcal{V}}

\newcommand{\head}{\mathsf{Head}}
\newcommand{\lin}{\text{lin}}

\usepackage{algorithm}
\usepackage{algpseudocode}
\renewcommand{\algorithmicrequire}{\textbf{Input:}}
\renewcommand{\algorithmicensure}{\textbf{Output:}}
\algrenewcomment[1]{\hfill {\color{OliveGreen!75} \(\triangleright\) #1}}
\algnewcommand{\LineComment}[1]{\State \(\triangleright\) #1}

\theoremstyle{plain}

\theoremstyle{definition}

\theoremstyle{remark}

\usepackage[textsize=tiny]{todonotes}

\icmltitlerunning{
  A Sober Look at LLMs for Bayesian Optimization Over Molecules
}

\begin{document}

\twocolumn[
  \icmltitle{
    A Sober Look at LLMs for Material Discovery: \\
    Are They Actually Good for Bayesian Optimization Over Molecules?
  }



  \icmlsetsymbol{equal}{*}

  \begin{icmlauthorlist}
    \icmlauthor{Agustinus Kristiadi}{vec}
    \icmlauthor{Felix Strieth-Kalthoff}{uoft}
    \icmlauthor{Marta Skreta}{uoft,vec}
    \icmlauthor{Pascal Poupart}{uw,vec}
    \icmlauthor{Al\'{a}n Aspuru-Guzik}{uoft,vec}
    \icmlauthor{Geoff Pleiss}{ubc,vec}
  \end{icmlauthorlist}

  \icmlaffiliation{vec}{Vector Institute}
  \icmlaffiliation{uoft}{University of Toronto}
  \icmlaffiliation{uw}{University of Waterloo}
  \icmlaffiliation{ubc}{University of British Columbia}

  \icmlcorrespondingauthor{Agustinus Kristiadi}{\href{mailto:akristiadi@vectorinstitute.ai}{akristiadi@vectorinstitute.ai}}

  \icmlkeywords{Machine Learning, ICML}

  \begin{center}
    {\footnotesize \bfseries \url{https://github.com/wiseodd/lapeft-bayesopt}}
  \end{center}

  \vskip 0.3in
]



\printAffiliationsAndNotice{}  

\begin{abstract}
  Automation is one of the cornerstones of contemporary material discovery. Bayesian optimization (BO) is an essential part of such workflows, enabling scientists to leverage prior domain knowledge into efficient exploration of a large molecular space. While such prior knowledge can take many forms, there has been significant fanfare around the ancillary scientific knowledge encapsulated in large language models (LLMs). However, existing work thus far has only explored LLMs for heuristic materials searches. Indeed, recent work obtains the uncertainty estimate---an integral part of BO---from point-estimated, \emph{non-Bayesian} LLMs. In this work, we study the question of whether LLMs are actually useful to accelerate principled \emph{Bayesian} optimization in the molecular space. We take a sober, dispassionate stance in answering this question. This is done by carefully (i) viewing LLMs as fixed feature extractors for standard but principled BO surrogate models and by (ii) leveraging parameter-efficient finetuning methods and Bayesian neural networks to obtain the posterior of the LLM surrogate. Our extensive experiments with real-world chemistry problems show that LLMs can be useful for BO over molecules, but only if they have been pretrained or finetuned with domain-specific data.
\end{abstract}

\vspace{-1.5em}



\section{Introduction}
\label{sec:intro}

\begin{figure}
  \centering

  \resizebox{\linewidth}{!}{%
    \includegraphics{figs/one.tikz}
  }

  \caption{
    LLMs seem to ``understand'' chemistry.
    However, they often produce completely wrong answers while sounding very convincing.
    Both APIs were accessed on 2024-01-07.
  }
  \label{fig:one}
\end{figure}

Material discovery describes the inherently laborious, iterative process of designing materials candidates, preparing them experimentally, testing their properties, and eventually updating the initial design hypothesis \citep{deregt2020, greenaway2023}.
While human researchers have largely driven this process for the last century, there is demand for more efficient automated methods in the face of pressing societal challenges related to health care, nutrition, or clean energy \citep{tom2024selfdrivingchem}.
Major challenges associated with the discovery process are the complex and black box-like mapping between a material's structure and its properties, as well as the vastness of the design space \citep{wang2023discovery}.

\begin{figure*}
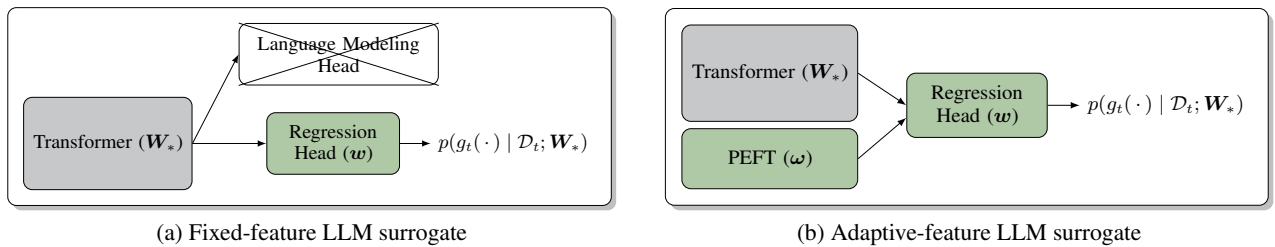

  \centering
  \subfloat[Fixed-feature LLM surrogate]{
    \resizebox{0.48\linewidth}{!}{%
      \includegraphics{figs/surrogate_fixed}
    }
  }
  \hfill
  \subfloat[Adaptive-feature LLM surrogate]{
    \resizebox{0.48\linewidth}{!}{%
      \includegraphics{figs/surrogate_peft.tikz}
    }
  }

  \caption{
    The surrogates we consider in this work.
    ``PEFT'' refers to parameter efficient finetuning which adds a (proportionally) few trainable weights \(\vomg\) to the transformer.
      {\color{Gray}Grey} denotes frozen weights that act as conditioning variables in the posterior over the surrogate \(g_t\).
      {\color{OliveGreen!80}Green} denotes weights that are inferred in a Bayesian manner (e.g., to obtain \(p(\vw, \vomg \mid \D_t)\)) and then marginalized over to obtain the posterior predictive distribution \(g_t\) (e.g., \(\iint p(g_t(\,\cdot\,) \mid \vw, \vomg; \mW_*) \, p(\vw, \vomg \mid \D_t) \, d\vw \, d\vomg\)).
    Both models are principled Bayesian surrogates, in contrast to the in-context learning frameworks considered by prior works on BO with LLMs \citep{ramos2023bayesopticl,anonymous2023llambo}.
  }
  \label{fig:surrogates}
  \vspace{0.5em}
\end{figure*}

To address the aforementioned problems, Bayesian optimization \citep[BO;][]{movckus1975bayesopt} has been increasingly used in chemistry \citep{gauche,hickman2023olympus2}.
Key components of successful BO include its \emph{priors} (informative priors imply efficient posterior inference with limited data) and its \emph{probabilistic surrogate models}---e.g.\ via Gaussian processes \citep{rasmussen2003gaussian,snoek2012practical} or Bayesian neural networks \citep{kim2022deep,li2023study,kristiadi2023promises}.
The probabilistic formulation of BO is useful since optimizing a black-box function is an inherently uncertain problem.
We do not know \emph{a priori} the form of the function and our approximation of it might be imprecise.
Probabilistic surrogate models are thus useful to quantify the inherent uncertainty surrounding the optimization landscape, allowing for principled approaches to the exploration-exploitation tradeoff \citep{garnett2023bayesopt}.

Good domain-specific priors,
a necessary component for reliable uncertainty estimates, are hard to define analytically.
Recent work has thus pursued \emph{implicit priors}
often obtained through pretrained feature extractors \citep{chithrananda2020chemberta,ross2022molformer}.
Large language models (LLMs)---which have become very popular in many domains that are traditionally rather disconnected from natural language processing such as biology \citep{vig2021bertologybiology}, education \citep{kasneci2023chatgpteducation}, law \citep{chalkidis2020legalbert}, and chemistry \citep[etc.]{jablonka2023,guo2023large,jablonka_schwaller_ortega-guerrero_smit_2023}---%
are one potential source of pretrained features for BO.
On the other hand, recent works have warned that LLMs might not necessarily understand natural language, but simply act as very expensive ``stochastic parrots'' \citep{bender2021stochasticparrots}; see \cref{fig:one}, for example.
Nevertheless, due to the apparent capabilities of LLMs, some recent works have leveraged off-the-shelf LLMs such as GPT-4 \citep{openai2023gpt4} for BO over molecules \citep{ramos2023bayesopticl} and hyperparameter tuning \citep{anonymous2023llambo}.
However, their uncertainty estimates are obtained only through heuristics, such as from the softmax probabilities of the generated answer tokens, coming from point-estimated non-Bayesian LLMs.
These non-Bayesian uncertainties thus might not be optimal for the exploration-exploitation tradeoff that is so crucial for BO \citep{garnett2023bayesopt}.

In this work, we take a dispassionate look at LLMs for BO over molecules.
We do so by carefully constructing and studying two kinds of surrogate models that are amenable to a principled Bayesian treatment (see \cref{fig:surrogates}).
First, we treat the LLM as a fixed feature extractor to test whether its pretrained embbeddings are already useful for BO over molecules.
Second, we measure to what degree the ``stochastic parrot'' can be ``taught''---via parameter-efficient fine-tuning methods (PEFT) \citep[e.g.,][]{houlsby2019adapter,li2021prefix,hu2022lora} and the Laplace approximation \citep{mackay1992evidence,daxberger2021laplace}---to perform efficient Bayesian exploration in the molecular space.

In sum, our contribution is four-fold:
\begin{enumerate}[(a)]
  \vspace{-0.85em}
  \setlength\itemsep{0em}
  \item We study the out-of-the-box usefulness of pretrained LLMs for material discovery by using their last-layer embeddings in BO.
  \item We study whether finetuning through PEFT and then applying approximate Bayesian inference over it is worth the effort in terms of the BO performance.
  \item We provide an easy-to-use software library for principled BO on discrete space with LLMs; see title page.\footnote{Supplementary experiment code can be found on {\footnotesize \url{https://github.com/wiseodd/llm-bayesopt-exps}}.}
  \item Through our extensive experiments (\(8\) real-world chemistry problems, \(8\) recent LLMs---including Llama-2---and non-LLM based features), we provide insights on whether, when, and how ``stochastic parrots'' can be useful to drive better scientific discovery.
        \vspace{-0.85em}
\end{enumerate}

\paragraph*{Limitations}
Our focus in this work is to study LLMs for discrete BO on a predetermined set of molecules, as usually done in real-world chemistry labs \citep[etc.]{laser2023}.
We leave the study of BO on continuous space with
LLM-based Bayesian surrogates as future work.
Finally, we focus only on chemistry, although our experiments can also be done for other domains.


\section{Preliminaries}
\label{sec:background}

Here, we introduce key concepts in Bayesian optimization, Bayesian neural networks, and large language models.

\subsection{Bayesian optimization}
\label{subsec:bo}

Suppose \(f: \X \to \Y\) is a function that is not analytically tractable and/or very expensive to evaluate.
We would like (without loss of generality) to find \(\vx_* = \argmax_{\vx \in \X} f(\vx)\).
For example, we might want to find a new drug \(\vx\) in the space of all drugs \(\X\) that has high efficacy over the population \(f(\vx)\).
An increasingly common way to approach this problem is to perform Bayesian optimization (BO).
The key components of BO are: (i) a \emph{surrogate function} \(g\) that tractably approximates \(f\); (ii) a prior belief and a likelihood (and hence a posterior) over \(g\);\footnote{In literature, they are often collapsed into a single notation \(p(f \mid \D)\). Here, we distinguish \(f\) and \(g\) for clarity.}
and (iii) an acquisition function \(\alpha: \X \to \R\) that implicitly defines a policy for choosing which \(\vx \in \X\) to evaluate \(f\) at.
The \emph{expressiveness} of \(g\) dictates how accurately we can approximate \(f\); and the \emph{calibration} of the
posterior (predictive) distribution \(p(g_t \mid \D_t)\) at step \(t\) under previous observations \(\D_t := \{ (\vx_i, f(\vx_i)) \}_{i=1}^{t-1}\) dictates where we should explore and where we should exploit in \(\X\).
This exploration-exploitation balance is the driving force behind the effectiveness of BO in finding the optimum \(\vx_*\) in a reasonable amount of time.

The \emph{de facto} choices of \(p(g_t \mid \D_t)\) are Gaussian processes \citep[GPs,][]{rasmussen2003gaussian}; although Bayesian neural networks (NNs) have also been increasingly used \citep{kim2022deep,kristiadi2023promises,li2023study}.
In the case of GPs, prior knowledge about the function \(f\) is injected into $g$ through the choice of prior covariance
(also known as the kernel function).
For NN-based surrogates, prior knowledge is determined through the choice of architecture \citep{kim2022deep}, a weight-space prior \citep{fortuin2022prior_revisited}, or through the usage of pretrained features \citep{rankovic2023bochemian}.
Finally, common choices for \(\alpha\) are expected improvement \citep[EI,][]{jones1998ei}, upper-confidence bound \citep[UCB,][]{auer2002ucb}, and Thompson sampling \citep[TS,][]{thompson1933sampling}.

{
\setlength{\textfloatsep}{0pt}
\begin{algorithm}[t]
  \small

  \caption{
    BO over a pool of molecules.
  }
  \label{alg:bo_molecules}

  \begin{algorithmic}[1]
    \Require
    \Statex Hard-to-evaluate function \(f\); surrogate function \(g\); candidate molecules \(\Dtest = \{ \vx_i \}_{i=1}^n\); initial dataset \(\D_1 = \{ (\vx_i, f(\vx_i)) \}_{i=1}^m\); time budget \(T\).

    \vspace{1em}

    \For {\(t = 1, \dots, T\)}      %
    \State \text{Compute posterior pred.} \(p(g_t \mid \D_t)\) \Comment{E.g.\ via GP or LLA}
    \State \(\vx_t = \argmax_{\vx \in \Dtest} \alpha(p(g_t(\vx) \mid \D_t))\)
    \State Compute \(f(\vx_t)\)
    \State \(\D_{t+1} = \D_t \cup \{ (\vx_t, f(\vx_t)) \}\)
    \State \(\Dtest = \Dtest \setminus \{ \vx_t \}\)
    \EndFor
    \State \Return \(\argmax_{(\vx, f(\vx)) \in \D_{T+1}} f(\vx)\)
  \end{algorithmic}
\end{algorithm}
}

\subsubsection{BO in chemical space}
\label{subsec:chem}

The discovery of molecular materials represents an optimization problem in a search space of discrete molecules that is estimated to contain at least $10^{100}$ unique molecules \citep{Restrepo2023}.
At the same time, the practical accessibility of this space is severely limited.
To date, only about $10^8$ molecules have been reported experimentally, and the synthesis of molecules has unanimously been described as the bottleneck of molecular materials discovery.
This applies to autonomous discovery in particular, where the limited robotic action space and the availability of reactants and reagents are additional constraints \citep{tom2024selfdrivingchem}.

Experimental discovery campaigns have usually constrained the search space to much smaller sets of accessible, synthesizable molecules.
Let \(\Dtest\) be such a set of candidate molecules.
The BO problem can then be treated as an optimization over a finite discrete set where \(\X = \Dtest\)---see \cref{alg:bo_molecules}.
Note that in this case, we do not need to perform continuous optimization (e.g., via SGD) to maximize the acquisition function \(\alpha\)---we can simply enumerate all molecules and pick the maximum.
While this approach can be expensive when \(\vert \Dtest \vert\) is large, it is parallelizable and easier than continuous optimization.

Nowadays, contract research organizations offer virtual, synthesizable libraries comprising billions of molecules, and offer synthesis-on-demand services \citep{EnamineReal, Gorgulla2023}.
In drug discovery, these libraries serve as the foundation for virtual screening efforts \citep{Shoichet2004, Schneider2010, PyzerKnapp2015, Lyu2019}, in which the library is sequentially filtered using progressively more costly computational tools, and the remaining candidates are evaluated experimentally.
BO has been used for finding optimal candidates in a virtual library, both with simulated \cite{Zhang2019, korovina2019chembo, gryffin, Hickman2022, gauche} and experimental \cite{laser2023, angello2023} objectives.

\subsection{Bayesian neural networks}
\label{subsec:bnns}

Let \(g: \X \times \varTheta \to \Y\) defined by \((\vx, \vtheta) \mapsto g_\vtheta(\vx)\) be a neural network (NN).
The main premise of Bayesian neural nets (BNNs) is to approximate the posterior \(p(\vtheta \mid \D)\) over the parameters of \(g\) via a simpler distribution that encodes uncertainty on \(\varTheta \subseteq \R^P\).
The standard point estimate:
\begin{equation} \label{eq:map}
  \vtheta_* = \argmax_{\vtheta \in \varTheta} \, \underbrace{\log p(\D \mid \vtheta) + \log p(\vtheta)}_{= \log p(\vtheta \mid \D) - \mathrm{const}} \, ,
\end{equation}
with the log-likelihood loss \(\log p(\D \mid \vtheta)\) and a regularizer \(\log p(\vtheta)\) over \(\vtheta\) can be seen as a Dirac distribution on \(\varTheta\).
However, it is \emph{not} a BNN since it has zero uncertainty according to any standard metric
(variance, entropy, etc.).

\subsubsection{Laplace approximations}
\label{subsubsec:laplace}

One of the simplest BNNs is the Laplace approximation \citep[LA,][]{mackay1992practical}, which has been increasingly used for BO \citep{kristiadi2023promises,li2023study}.
Given a (local) maximum \(\vtheta_*\), the LA fits a Gaussian \(q(\vtheta \mid \D) := \N(\vtheta_*, \mSigma_*)\) centered at \(\vtheta_*\) with covariance given by the inverse-Hessian \(\mSigma_* = (-\nabla^2_\vtheta \log p(\vtheta \mid \D) \vert_{\vtheta_*})^\inv\).

A popular instantiation of the LA is the linearized Laplace approximation \citep[LLA][]{immer2021improving}, which approximates the Hessian via the Gauss-Newton matrix \citep{botev2017practical} and performs a linearization \(g^\lin_\vtheta(\vx) = g_{\vtheta_*}(\vx) + \mJ_*(\vx) \cdot (\vtheta - \vtheta_*)\) of the NN over \(\vtheta\).
Here, \(\mJ_*(\vx)\) is the Jacobian matrix \((\partial g / \partial \vtheta \vert_{\vtheta_*})\) of the network at \(\vtheta_*\).
Note that, the network function \(\vx \mapsto g^\lin_\vtheta(\vx)\) is still \emph{non}-linear.
Crucially, due to the linearity of \(g\) over \(\vtheta\) and the Gaussianity of \(\vtheta\), the output/predictive distribution \(p(g^\lin(\vx) \mid \D) = \int g^\lin_\vtheta(\vx) \, q(\vtheta \mid \D) \, d\vtheta\) is also Gaussian, given by
\begin{equation} \label{eq:lla_marginal}
  p(g^\lin(\vx) \mid \D) = \N\left(g_{\vtheta_*}(\vx), \mJ_*(\vx) \mSigma \mJ_*(\vx)^\top\right) .
\end{equation}
In fact, \(p(g^\lin \mid \D)\) is a GP with a mean function given by the NN \(g\) and a covariance function that is connected to the empirical neural tangent kernel \citep{jacot2018ntk}.
These facts make the LLA intuitive yet powerful: it adds an uncertainty estimate to the original NN prediction \(g_{\vtheta_*}(\vx)\).\footnote{Non-Gaussian output distributions can also be constructed, see e.g.\ \citep{kristiadi2022refinement,bergamin2023riemannian}.}

Furthermore, the hyperparameters that dictate the NN prior (which we denote as $\vgamma$) can be tuned
via the LA's marginal-likelihood approximation
\citep{daxberger2021laplace}:
\begin{equation} \label{eq:lla_marglik}
  \resizebox{0.89\hsize}{!}{
    \(\textstyle Z(\vgamma) = \log p(\vtheta_* | \D; \vgamma) + \frac{P}{2} \log 2\pi + \frac{1}{2} \log \abs{\mSigma_*(\vgamma)}\)
  } ,
\end{equation}
where we have made the dependency of the posterior and the Hessian on the hyperparameters \(\vgamma\) explicit.
For example, \(\vgamma\) could
contain the weight decay strength (corresponding to the prior precision of the Gaussian prior on \(\varTheta\)) as well as the noise strength in the likelihood of \(g\).

\subsection{Large language models}
\label{subsec:llms}

A crucial component of the recent development in large NNs is the \emph{\(K\)-head self-attention} mechanism \citep{vaswani2017attention}.
Given a length-\(T\) sequence of input embeddings of dimension \(N\), say \(\mX \in \R^{T \times N}\), it computes
\begin{equation}
  \begin{aligned}
    \textstyle
    \mO   & = [\mH_1, \dots, \mH_K] \mW_o^\top \in \R^{T \times O} ,                                                                       \\
    \mH_i & = s \left( \textstyle{\frac{1}{\sqrt{D}}} (\mX \mQ_i^\top) (\mX \mK_i^\top)^\top \right) (\mX^\top \mV_i) \in \R^{T \times D},
  \end{aligned}
\end{equation}
where \([\dots]\) is a column-wise stacking operator, taking \(K\)-many \(T \times D\) matrices to a \(T \times KD\) matrix; \(\smash \mW_o \in \R^{O \times KD}\) and \(\smash \mQ_i, \mK_i, \mV_i \in \R^{D \times N}\) are linear projectors; and the softmax function \(s(\cdot)\) is applied row-wise.

The resulting network architecture, obtained by stacking multiple attention modules along with other layers like residual and normalization layers, is called a \emph{transformer}.
When used for language modeling, the resulting model is called a \emph{large language model (LLM)}.
The output \(\mO\) of the last transformer module can then be used as a feature for a dense output layer \(\head: \R^O \to \R^C\), taking the row-wise aggregate (e.g.\ average) of \(\mO\) to a \(C\)-dimensional vector, where \(C\) is the number of outputs in the problem.
For natural language generation, \(C\) equals the size of the vocabulary \(\V\), e.g.\ around \num{32000} in \citet{touvron2023llama}.
One can also modularly replace this head so that the LLM can be used for different tasks, e.g. single-output regression where \(C = 1\).

\subsubsection{Parameter-Efficient Fine-Tuning}
\label{subsubsec:peft}

Due to their sheer size, the cost of training LLMs from scratch is prohibitively expensive even for relatively small models \citep{sharir2020cost}.
Thankfully, LLMs are usually trained in a task-agnostic manner and have been shown to be meaningful, generic ``priors'' for natural-language-related tasks \citep{brown2020gpt3}.
One can simply finetune a pretrained LLM to obtain a domain-specific model \citep{sun2019bert_sentiment}.
However, standard finetuning, i.e.\ further optimizing \emph{all} the LLM's parameters, is expensive.
Parameter-efficient fine-tuning (PEFT) methods---which add a few additional parameters \(\vomg\) to the LLM and keep the original LLM parameters frozen---have therefore become standard.

\begin{figure*}[t!]
  \centering
  \includegraphics[width=\linewidth]{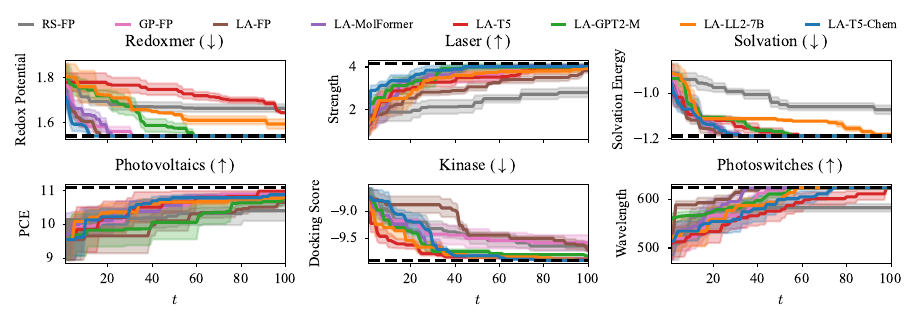}

  \vspace{-1em}
  \caption{
    LLM as fixed feature extractors in BO over molecules.
    See \cref{fig:surrogates}(a) for the model schematic.
  }
  \label{fig:last_layer_bo}
\end{figure*}

A popular example of PEFT is LoRA \citep{hu2022lora}, which uses a bottleneck architecture to introduce additional parameters in a LLM.
Let \(\mW_* \in \R^{D \times N}\) be an attention weight matrix.
LoRA freezes \(\mW_*\) and augments it into
\begin{equation}
  \mW = \mW_* + \mB^\top \mA; \qquad \mA \in \R^{Z \times N}, \mB \in \R^{Z \times D} .
\end{equation}
If \(Z\) is relatively small, the matrices \(\mA, \mB\) will introduce just a few additional parameters.
Note that, many other PEFT methods are also commonly used in practice, e.g., Adapter \citep{houlsby2019adapter}, Prefix Tuning \citep{li2021prefix}, IA3 \citep{liu2022ia3}, etc.
That is LoRA, is not the only choice for performing PEFT.


\section{Experiment Setup}
\label{sec:setup}

Equipped with the necessary background knowledge from \cref{sec:background}, we now discuss our experiments
to test whether LLMs are good for BO in molecular discovery.
We refer the reader to \cref{alg:bo_molecules} for the concrete problem statement.

\textbf{Datasets}\enspace
We evaluate the models considered (see below) on the following datasets that represent realistic problem sets from molecular materials discovery:
(i) minimizing the redox potential (\textbf{redoxmer}) and (ii) minimizing the solvation energy (\textbf{solvation}) of possible flow battery electrolytes \citep{agarwal2021}, (iii) minimizing the docking score of \textbf{kinase} inhibitors for drug discovery \citep{graff2021}, (iv) maximizing the fluorescence oscillator strength of \textbf{lasers} \citep{laser2023}, (v) maximizing the power conversion efficiency (PCE) of \textbf{photovoltaics} materials \citep{lopez2016cleanenergy}, and (vi) maximizing the \(\pi\)-\(\pi^*\) transition wavelength of organic \textbf{photoswitches} \citep{griffiths2022photoswitches}.
These problems cover a variety of molecular physical properties and therefore represent a diverse set of molecular design tasks.
For each virtual library of molecules above,
we use physics-inspired simulators proposed by the respective authors as the ground truth functions \(f(\vx)\).\footnote{We refer the reader to \cref{app:subsec:datasets} for further details regarding all the datasets we use.}

\textbf{Features and LLMs}\enspace
As non-LLM baselines, we use
\(1024\)-bit Morgan \textbf{fingerprints} \citep{morgan1965fingerprints} as a chemistry-specific algorithmic vectorization scheme, and the feature vectors from the pretrained \textbf{MolFormer} transformer \citep{ross2022molformer}.
Meanwhile, for the general-purpose LLMs, we use various recent architectures of varying sizes:
T5-Base \citep[\textbf{T5},][]{raffel2020t5}, GPT-2-Medium \citep[\textbf{GPT2-M},][]{radford2019gpt2}, and Llama-2-7b \citep[\textbf{LL2-7B},][]{touvron2023llama}.
Finally, we use the work of \citet[\textbf{T5-Chem}]{christofidellis2023unifying} to represent domain-specific LLMs.

\vspace{0.5em}

\textbf{Prompts}\enspace
For text-based surrogate models, we consider several prompting functions \(c(\vx)\) that map molecules \(\vx\) to sentences.
They are (i) \textbf{just-smiles} which contains just the SMILES \citep{weininger1988smiles} representation of \(\vx\), (ii) \textbf{completion} which treats the predicted \(f(\vx)\) as a completion to the sentence, (iii) \textbf{naive} which asks the LLM for \(f(\vx)\), and (iv) \textbf{single-number} which augments \emph{naive} with an additional prompt to the LLM to only output numbers.
(Details in \cref{app:subsec:prompts}.)
Unless specified explicitly, the default prompt we use is \emph{just-smiles}.

\vspace{0.5em}

\textbf{Evaluation}\enspace
In addition to measuring BO performance via the problem-specific optimum values over time, we use the GAP metric \citep{jiang2020binoculars} which provides a normalized (i.e., problem-independent) counterpart.
The GAP metric is useful since it allows us to compare and aggregate performance across datasets.
For multiobjective BO experiments, we measure performance using the standard hypervolume metric \citep{zitzler1999evolutionary} which computes the volume of the current Pareto front found by the surrogate.


\section{How Informative are Pretrained LLMs?}
\label{sec:llm_feat}

First, we study the out-of-the-box, non-finetuned capability of LLMs for BO.
To this end, we treat an LLM as a \emph{fixed} feature extractor:
Given a pretrained LLM, we remove its language-modeling head and obtain the function \(\phi_{\mW_*}\), mapping a textual context \(c(\vx)\) of a molecule \(\vx\) into its final transformer embedding vector \(\phi_{\mW_*}(c(\vx)) \in \smash{\R^H}\).
We can then apply a standard surrogate model \(g_\vtheta: \R^H \to \R\) like GPs or BNNs on \(\smash{\R^H}\).
See \cref{fig:surrogates}(a) for the illustration and \cref{alg:llm_feat_bo} in \cref{app:sec:details} for the BO loop.

We use two commonly-used surrogate models over the fixed LLM and non-LLM features: (i) a GP with the Tanimoto and the Mat\'{e}rn kernels for the fingerprints and LLM/MolFormer features, respectively \citep{gauche}, and (ii) a Laplace-approximated \(3\)-layer ReLU NN with \(50\) hidden units on each layer, following the finding of \citet{li2023study}.
The Thompson sampling acquisition function is used in all experiments due to its simplicty and increasing ubiquity
in chemistry applications \citep{hernandez2017parallelTS}.
Refer to \cref{app:subsec:acqfs} for results with the expected improvement acquisition function.

\subsection{General or domain-specific LLMs?}
\label{subsec:general_vs_chem}

\begin{figure}
  \begin{center}
    \includegraphics[width=\linewidth]{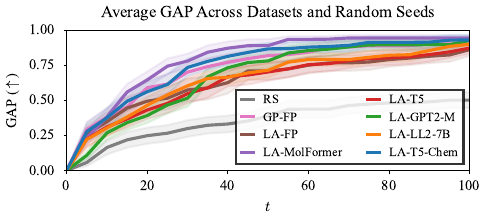}
  \end{center}

  \vspace{-1em}

  \caption{
    Summarized performance of the results in \cref{fig:last_layer_bo} in terms of the GAP metric.
    Chemistry-focused features (T5-Chem, MolFormer, and even fingerprints) are better than
    general-purposed LLM features.
  }
  \label{fig:gap_fixed}
\end{figure}

We present our first set of results in \cref{fig:last_layer_bo,fig:gap_fixed} (the latter figure summarizes the former).
First, \lafp{} surrogates are competitive with or better than \gpfp{} surrogates on the majority of the problems when using fingerprint features.
Thus we only consider LA surrogates for LLM features.
See also \cref{fig:gap_gp} in \cref{app:subsec:la_vs_gp} for additional comparisons between the LA and GP which further support our decision in mainly using the LA for the rest of our experiments.

We note that features obtained from general-purpose LLMs (\latfive{}, \lagpt{}, and \lallama{}) tend to underperform the simple fingerprints baseline.
This indicates that although general-purpose LLMs seem to ``understand'' chemistry as illustrated in \cref{fig:one}, the \emph{features} encoded by these LLMs are less informative for chemistry-focused BO.
While this conclusion holds in our specific problem setup,
we note that other studies find LLMs to be useful for more general problems \citep{gruver2023large,han2023context}.

Meanwhile, chemistry-specific transformer features (\latfivechem{}, \lamf{}), are generally better-suited than the general-purpose LLM features.
However, we note that \latfivechem{} LLM features perform slightly worse on average than the non-LLM \lamf{} features.
Considering that \latfivechem{} is larger than \lamf{} (\(220\)M vs.\ \(44\)M parameters) and that \lamf{} is trained using more chemistry data (\(100\)M vs.\ \(33\)M), this finding may indicate that domain-specific pretraining data matters more than the natural language capability of a transformer model.

\begin{tcolorbox}[colback=white, colframe=black, rounded corners, left=5pt, right=5pt, top=5pt, bottom=5pt, boxsep=5pt, boxrule=0.75pt]
  Domain-specific transformers are useful as feature extractors in BO over molecules.
  They tend to outperform general-purpose LLMs and traditional fingerprint features.
  However, this may be due to the transformer's capacity and the chemistry-specific pretraining data, not so much its natural language capability.
\end{tcolorbox}

\subsection{Multiobjective optimization}
\label{sec:multiobj}

In addition to the single-objective problems in \cref{sec:llm_feat,sec:lapeft}, we perform multiobjective BO experiments by (i) combining both objectives in the flow battery problem above, and (ii) adding an extra maximization objective (electronic gap) to the laser problem.
We refer to these problems as \textbf{multi-redox} and \textbf{multi-laser}, respectively.

To accommodate the additional objectives, we cast the problems as multi-output regression problems---for each \(\vx\), the posterior of \(g(\vx)\) is thus a multivariate Gaussian over \(\R^C\) where \(C\) is the number of the objectives.
For the acquisition function, we use the scalarized Thompson sampling \citep{paria2020flexible} with a fixed, uniform weighting.

\begin{figure}[t]
  \centering
  \includegraphics[width=\linewidth]{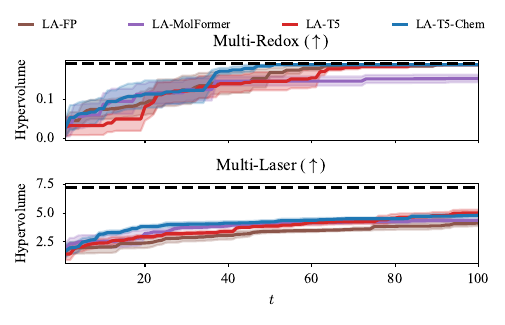}

  \vspace{-1em}
  \caption{
    Multiobjective BO performance in terms of the standard hypervolume metric.
  }
  \label{fig:multiobj_scalarized}
\end{figure}

The results, in terms of the standard hypervolume (under the estimated Pareto frontier) evaluation metric \citep{zitzler1999evolutionary}, are provided in \cref{fig:multiobj_scalarized}.
We found that \latfivechem{} performs best overall.
When \(t\) is small, the non-LLM chemistry-specific transformer \lamf{} is better than \latfive{} and \lafp{}.
However, \lamf{} underperforms both LLMs at the latter stages of the optimization.
We hypothesize that the smaller model size of \lamf{} (see the preceding section) might contribute to this underperformance.
In any case, our conclusion here is consistent with
the previous section.

\subsection{Effects of prompting}
\label{subsec:prompting}

Here we test how prompting affects BO performance,
comparing the prompts described in \cref{sec:setup}.
We present the results for the redoxmer and photoswitches tasks in \cref{fig:prompts_assorted}
(see \cref{fig:prompts_full} in the appendix for the rest of the problems).

\begin{figure}[h!]
  \centering
  \includegraphics[width=\linewidth]{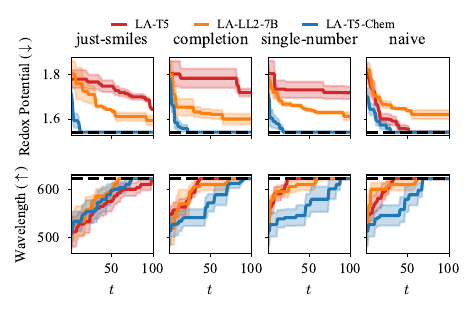}

  \vspace{-1em}
  \caption{
    BO results across prompts.
    \textbf{Top:} Redoxmer, \textbf{bottom:} Photoswitches.
    Results for the other datasets are in \cref{app:subsec:prompts}.
  }
  \label{fig:prompts_assorted}
\end{figure}

Prompting does indeed make a difference: unlike general LLMs (\latfive{}, \lallama{}), the chemistry-specific \latfivechem{} works best when the prompt is simply the SMILES string itself.
Nevertheless, we note that \latfivechem{} obtains the best performance in most of the problems considered and across all prompts---see both \cref{fig:prompts_assorted,fig:prompts_full}.
Thus the chemistry-specific \latfivechem{} both yield better BO performance while not requiring prompt engineering.

In \cref{fig:smiles_vs_iupac,fig:iupac_prompts} (\cref{app:sec:results}), we show results with IUPAC representation of molecules instead of SMILES.
Note that IUPAC strings are closer to natural language than SMILES.
E.g., the molecule \ H\(_2\)SO\(_4\) has IUPAC name ``sulfuric acid'' and SMILES representation ``OS(=O)(=O)O''.
We draw a similar conclusion as in the preceding section that the choice of which  representation to use is LLM-dependent.
For \latfivechem{}, SMILES is preferable, consistent with how it was pretrained \citep{christofidellis2023unifying}.

\begin{tcolorbox}[colback=white, colframe=black, rounded corners, left=5pt, right=5pt, top=5pt, bottom=5pt, boxsep=5pt, boxrule=0.75pt]
  Prompting does impact BO performance.
  It is preferable to stick with a prompt that is close to the one used for pretraining the LLM.
\end{tcolorbox}

\subsection{The case of in-context learning}
\label{subsec:icl_baseline}

Finally, we compare Laplace-approximated surrogate models
against the recently proposed in-context learning (ICL) optimizer method of \citet[\textbf{BO-LIFT},][]{ramos2023bayesopticl}.
BO-LIFT works purely by prompting chat-based models such as GPT-4 \citep{openai2023gpt4} and Llama-2-7b \citep[the chat version]{touvron2023llama2}.
See \cref{app:subsec:incontext_baselines} for details.

We note that the uncertainty estimates yielded by BO-LIFT are obtained based on the variability in the decoding steps of the LLM.
They are thus not Bayesian since they still arise from a point-estimated model.
In contrast, all the Bayesian surrogates we consider in this work approximate the posterior distribution over the LLM's weights.

\begin{figure}[h!]
  \centering
  \includegraphics[width=\linewidth]{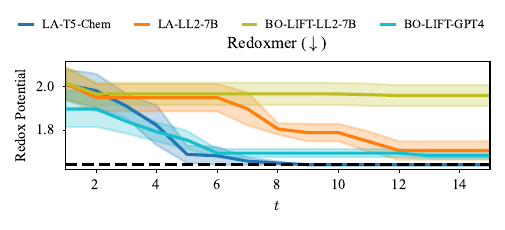}

  \vspace{-1em}
  \caption{
    Fixed-feature BO surrogates vs.\ the in-context-learning optimizer of \citep{ramos2023bayesopticl} on the Redoxmer dataset.
  }
  \label{fig:icl_baseline}
\end{figure}

We present the result of the subsampled Redoxmer dataset in \cref{fig:icl_baseline} (\(\vert\Dtest\vert = 200, \vert\D_1\vert = 5, T = 15\) using the notation of \cref{alg:bo_molecules}).
We subsample due to the cost of the experiment, as discussed below.
We find that BO-LIFT is ineffective when combined with \boliftllama{}.
Meanwhile, it performs much better with \boliftgpt{}, indicating that ICL may require a very large, expensive LLM.
Indeed, each optimization run costs between \$\num{12}-\$\num{18} USD for \boliftgpt{}, totaling to \$\num{75.81} USD over \num{5} random seeds.

In contrast, using a chemistry-specific, small (\num{200}M parameters) \latfivechem{} as a feature extractor for a principled BO surrogate is better \emph{and} much cheaper.
Indeed, \latfivechem{} can be run on even mid-range consumer-grade GPUs and the LA or GP surrogate's training can be done on CPUs.

\begin{tcolorbox}[colback=white, colframe=black, rounded corners, left=5pt, right=5pt, top=5pt, bottom=5pt, boxsep=5pt, boxrule=0.75pt]
  A chemistry-specific LLM combined with a principled Bayesian surrogate is preferable to an ICL optimizer, both in terms of performance and cost.
\end{tcolorbox}



\section{How Useful are Finetuned LLMs?}
\label{sec:lapeft}

In the previous section, we have seen that we can use an LLM as a \emph{fixed} feature extractor in a BO loop with standard surrogate models.
Here, we answer the question of whether or not treating the whole LLM itself as the surrogate model improves BO.
The hypothesis is that we can improve BO performance by performing feature learning---adapting the LLM feature to the problem at hand.

\begin{figure*}
  \centering
  \includegraphics[width=\textwidth]{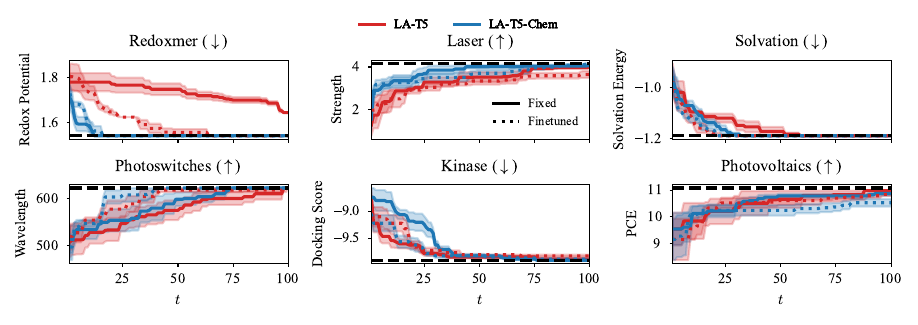}

  \vspace{-1em}
  \caption{
    Finetuned LLMs as surrogate models compared against the fixed-feature surrogates from \cref{sec:llm_feat}.
  }
  \label{fig:fine_tuning}
\end{figure*}

How should we compute the posterior \(p(g_t \mid \D_t)\) of a LLM with learned features?
Let \(g_\vtheta = \head_{\vw} \circ \varphi_{\mW}\)
be an LLM feature extractor with weights \(\mW\) composed with a regression head \(\head_\vw\) with weights \(\vw\).
Here, \(\vtheta = \{ \mW, \vw \}\).
Given the dataset \(\D_t\) at time \(t\), the seemingly most straightforward way to perform finetuning on \(g_\vtheta\) and obtaining its posterior is to perform a Bayesian update on the previous posterior of \(\mW\) using new observations in \(\D_t\) \citep{shwartz2022bayesian_transfer}.
However, this requires an existing posterior over \(\mW\) in the first place, which can be \emph{very} costly to obtain.

Recall that a more tractable way to introduce feature learning
is to leverage PEFT.
However, unlike full finetuning, it does not admit a straightforward Bayesian posterior update on \(\mW\).
In particular, how should one incorporate the pretrained LLM weights \(\mW_*\) into the Bayesian inference over the PEFT weights?
Here, we generalize the work of \citet{yang2023loralaplace}---which applies the LA on LoRA weights---to make it compatible with any general PEFT method.

We treat the original LLM weights \(\mW\) as hyperparameters and perform Bayesian inference only on \(\vw\) and the PEFT weights \(\vomg\).
Specifically, let \(g_\vtheta = \head_{\vw} \circ \varphi_{\vomg, \mW_*}\) with \(\vtheta = \{ \vw, \vomg \}\) be the new surrogate model, and \(\mW_*\) be the pretrained LLM weights.
We define the posterior:\footnote{The dependence of the prior \(p(\vtheta; \mW = \mW_*)\) on \(\mW\) is useful, e.g.\ for initialization \citep{li2021prefix}.}
\begin{multline}
  \label{eq:prob_peft}
  p(\vtheta \mid \D; \mW = \mW_*) \propto p(\vtheta; \mW = \mW_*) \\
  \times p(\D \mid \vtheta; \mW = \mW_*) .
\end{multline}
In other words, the pretraining weights \(\mW_*\) act as \emph{conditioning variables} on the PEFT, just like any other hyperparameters.
The usual training procedure, i.e.\ finding the optimal PEFT weights \(\vtheta_*\), can then be seen as MAP estimation under this probabilistic framework.
While this formulation is rather obvious after the fact, so far, it has not been clarified in previous work.
Indeed, a clear interpretation of Bayesian PEFT in relation to Bayesian updating in full finetuning has been missing, even from the work of \citet{yang2023loralaplace}.

Notice that any Bayesian method can be used to approximate the posterior in \eqref{eq:prob_peft}.
In this work, we use the LA:
Given the weight-space LA posterior \(p(\vtheta \mid \D; \mW = \mW_*)\) over the PEFT weights, we can obtain the PEFT posterior predictive \(p(g_t \mid \D_t; \mW_*)\) via the LLA \eqref{eq:lla_marginal}.
This step is tractable since the Jacobian matrix \(\smash{\mJ_*}(\vx)\) is only of size \(C \times \smash{\widehat{P}}\), where \(C\) is the number of BO objectives (\emph{much} fewer than the language-modeling head's outputs) and \(\smash{\widehat{P}}\) is the PEFT parameters (\emph{much} fewer than the LLM's parameters; often less than \(1\%\) \citep{hu2022lora}).

The formulation of \eqref{eq:prob_peft} highlights an additional benefit of the LA:
since \(\mW\) is now a hyperparameter (i.e.\ it is part of \(\gamma\) in \eqref{eq:lla_marglik}) one can further optimize it (or a subset of it) via the marginal likelihood.
This optimization can potentially improve performance,\footnote{Notice the dependence of the likelihood on \(\mW\) in \eqref{eq:lla_marglik}.} similar to deep kernel learning \citep{wilson2016deep} in the context of GPs.
However, it also makes PEFT training as expensive as full finetuning.
We leave optimization of \(\mW\) for future work since our present focus is on the usage of LLMs in BO and not on new methods arising from the probabilistic model in \eqref{eq:prob_peft}.

\subsection{Are finetuned LLM surrogates preferable?}
\label{subsec:finetuning_experiments}

Following \citep{yang2023loralaplace}, we use LoRA as the PEFT method of choice:
For each time \(t\), we reinitialize and train the LoRA weights with MAP estimation \eqref{eq:map} using the observed molecules \(\D_t\), and then apply the LLA to obtain the posterior predictive distribution \(p(g_t \mid \D_t)\).
See \cref{fig:surrogates}(b) and \cref{alg:lapeft} for an illustration and pseudocode, respectively.
See also \cref{app:subsec:trainings} for the training details.
We compare the resulting finetuned surrogates with their fixed-feature surrogate counterparts from \cref{sec:llm_feat}.

\begin{figure}
  \begin{center}
    \includegraphics[width=\linewidth]{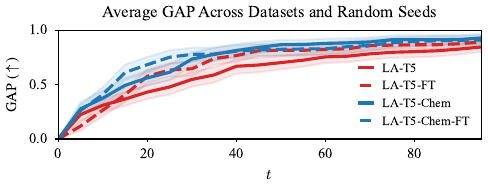}
  \end{center}

  \vspace{-1em}

  \caption{
    Summarized performance of the finetuning results in \cref{fig:fine_tuning} in terms of the GAP metric.
    Finetuning does seem to be beneficial for BO performance.
  }
  \label{fig:gap_ft}
  \vspace{-1em}
\end{figure}

In \cref{fig:fine_tuning,fig:gap_ft}, we show the finetuning results on \latfive{} and \latfivechem{}, representing general-purpose and chemistry-specific LLMs, respectively.
We found that finetuning is indeed beneficial for both cases---notice that it improves the BO performance in most problems compared to the fixed-feature version.

On the flip side, for some tasks, finetuning does not offer significant improvement over fixed-feature LLMs.
Moreover, in one problem (Photovoltaics), we found that finetuning decreases the performance of \latfivechem{}.
We attribute this degradation to the fact that we use the same hyperparameters (learning rate, weight decay, etc. of LoRA's SGD and the LA) on all problems.
This setup mirrors common practice: it is standard to simply use the default hyperparameters of a BO algorithm provided by a software package such as BoTorch \citep{balandat2020botorch}.
In any case, it is encouraging to see that finetuning generally works well across most BO problems,
even when only considering default hyperparameters.

The cost of each BO iteration \(t\) is largely bottlenecked by forward passes over the candidate molecules in \(\Dtest\) (\cref{alg:bo_molecules}, line 3), and \emph{not} the finetuning and Laplace approximation of the surrogate (see \cref{fig:timing} in \cref{app:sec:results}).
This is because \(\vert \Dtest \vert\) can be several orders of magnitude larger than \(\vert \D_t \vert\) during the BO loop, and forward passes on LLMs are generally expensive.
Additionally, the LLA posterior \eqref{eq:lla_marginal} requires computing the Jacobian of the network, amounting to the cost of several backward passes, depending on \(C\).
Due to GPU memory limitation, we can only use a minibatch size of \num{16}.
We note that fixed-feature surrogates do not have this problem since one can simply cache the LLM's features to be used for all iterations \(t\).
Nevertheless, this bottleneck can be alleviated via engineering efforts such as parallelizing the forward passes over \(\Dtest\).

\begin{tcolorbox}[colback=white, colframe=black, rounded corners, left=5pt, right=5pt, top=5pt, bottom=5pt, boxsep=5pt, boxrule=0.75pt]
  Finetuned LLM surrogates are preferable to their fixed-feature counterparts for both general and domain-specific LLMs.
  The bottleneck associated with finetuning is forward passes over \(\Dtest\) rather than training.
\end{tcolorbox}


\section{Related Work}
\label{sec:related}

While LLMs have been leveraged for BO \citep{ramos2023bayesopticl,anonymous2023llambo}, so far they have only been used in a heuristic manner:
uncertainty estimates are obtained from the softmax probabilities outputs of point-estimated (i.e., non-Bayesian) LLMs.
Meanwhile, \citet{ai4science2023impact} study the optimization capability of GPT-4 in a pure prompting scheme without uncertainty estimation of any kind.

The closest work to ours is that of \citet{rankovic2023bochemian}, who study the usage of text embeddings for BO with GP surrogates.
However, the goal of their work differs from ours:
We investigate whether the usage of LLMs is justified for BO due to their apparent chemistry question-answering capabilities
while \citep{rankovic2023bochemian} primarily study the usage of text embeddings in BO.
Furthermore, they do not study the effect of prompting and finetuning.

The present work can be seen as an extension to the LoRA LA work of \citet{yang2023loralaplace}.
Our work provides a clear probabilistic interpretation of their method, while also generalizing it and using it in the context of BO for chemistry.

Beyond discrete-set optimization, one can cast BO over molecules as a continuous optimization problem with the help of generative models.
\citet{gomez2018automatic,tripp2020sample,maus2022localbo} employ variational autoencoders to construct continuous latent representations of molecules and perform BO in these latent spaces.
The surrogate model and the autoencoder can be trained jointly \citep{stanton2022denoisingbo,maus2022localbo},
echoing our finetuning approach in \cref{subsec:finetuning_experiments}.
Unlike these prior works, we focus on studying the role of LLMs in the simpler yet practically relevant setting of discrete-set BO.


\section{Conclusion}
\label{sec:conclusion}

We have shown that large language models (LLMs) do indeed carry useful information to aid Bayesian optimization (BO) over molecules.
However, their usefulness is only apparent when one (i) uses a chemistry-specific LLM or (ii) performs finetuning---preferably both.
Indeed, for point (i), even when the recent Llama-2-7b LLM was used as a feature extractor for a BO surrogate or when the state-of-the-art GPT-4 was used in conjunction with in-context learning, the optimization performance was subpar compared to that of a \emph{much} smaller chemistry-focused LLM.
We address point (ii) by providing a general way to formulate Bayesian inference for parameter-efficient finetuning (PEFT) methods, which in turn enables principled uncertainty estimation over LLMs with any PEFT method.
We find that these principled BO surrogates are effective and yet much cheaper than in-context learning methods since small domain-specific LLMs can be used.
We hope that our findings and the accompanying software library can be useful for practitioners and inspire future principled methods around LLMs for scientific discovery, both inside and outside of chemistry.
In future work, we aim to characterize the mechanisms underpinning how LLMs induce a capable exploration-exploitation tradeoff.


\section*{Impact Statement}

This paper presents work whose goal is to advance the field of machine learning for chemistry.
We acknowledge the potential dual use of chemistry-specific models (and large language models in general) to search for materials for nefarious purposes.
Although the search spaces presented in this work are, to the best of our knowledge, highly unlikely to yield such materials, we recognize the necessity for safeguards in such efforts and encourage open discussions about their development.

\section*{Acknowledgments}

Resources used in preparing this research were provided, in part, by the Province of Ontario, the Government of Canada through CIFAR, and companies sponsoring the Vector Institute.
AK thanks Runa Eschenhagen and Alexander Immer for helpful discussions.
FSK is a postdoctoral fellow in the Eric and Wendy Schmidt AI in Science Postdoctoral Fellowship Program, a program by Schmidt Futures.

\bibliography{main}
\bibliographystyle{icml2024}

\newpage

\begin{appendices}
  \crefalias{section}{appendix}
  \crefalias{subsection}{appendix}
  \crefalias{subsubsection}{appendix}

  \onecolumn

  {
    \centering
    \Large
    \vspace{3em}
    \textbf{Appendices: A Sober Look at LLMs for Material Discovery} \par
    \vspace{1em}
  }

  \thispagestyle{plain}


\section{Additional Details}
\label{app:sec:details}

\subsection{Pseudocodes}
\label{app:subsec:pseudocodes}

We present the pseudocode of the BO loop corresponding to \cref{fig:surrogates}(a) and \cref{sec:llm_feat} in \cref{alg:llm_feat_bo}.
Meanwhile, the pseudocode corresponding to \cref{fig:surrogates}(b) and \cref{sec:lapeft} is in \cref{alg:lapeft}.

\begin{algorithm}[h!]
  \small

  \caption{
    Using an LLM as a feature extractor in BO.
  }
  \label{alg:llm_feat_bo}

  \begin{algorithmic}[1]
    \Require
    \Statex Pre-trained LLM feature extractor \(\phi_{\mW_*}\), mapping a context \(c(\vx)\) to its last transformer embedding vector \(\vh \in \smash{\R^H}\); prompting function \(c\) mapping a molecule \(\vx\) to a textual prompt; surrogate model \(g_\vtheta: \smash{\R^H} \to \R\); candidate molecules \(\Dtest = \{ \vx_i \}_{i=1}^n\); initial dataset \(\D_1 = \{ (\vx_i, f(\vx_i)) \}_{i=1}^m\); time budget \(T\).

    \vspace{0.5em}

    \For {\(t = 1, \dots, T\)}
    \State \(\varPhi_t = \{ (\phi_{\mW_*}(c(\vx)), f(\vx)) : (\vx, f(\vx)) \in \D_t \}\)
    \State \(p(g_t \mid \D_t) = \mathtt{infer}(g_\vtheta, \varPhi_t)\) \Comment{E.g.\ via GP or the LLA}
    \State \(\vx_t = \argmax_{\vx \in \Dtest} \alpha(p(g_t(c(\vx)) \mid \D_t))\)
    \State \(\D_{t+1} = \D_t \cup \{ (\vx_t, f(\vx_t)) \}\)
    \State \(\Dtest = \Dtest \setminus \{ \vx_t \}\)
    \EndFor
    \State \Return \(\argmax_{(\vx, f(\vx)) \in \D_{T+1}} f(\vx)\)
  \end{algorithmic}
\end{algorithm}

\begin{algorithm}[h!]
  \small

  \caption{
    LLM and PEFT as a surrogate model in BO.
  }
  \label{alg:lapeft}

  \begin{algorithmic}[1]
    \Require
    \Statex An LLM with a regression head \(g_\vtheta\); prompting function \(c\) mapping a molecule \(\vx\) to a textual prompt; candidate molecules \(\Dtest = \{ \vx_i \}_{i=1}^n\); initial dataset \(\D_1 = \{ (\vx_i, f(\vx_i)) \}_{i=1}^m\); time budget \(T\).

    \vspace{0.5em}

    \For {\(t = 1, \dots, T\)}
    \State \(\mathcal{C}_t = \{ (c(\vx), f(\vx)) : (\vx, f(\vx)) \in \D_t \}\)
    \State \(p(\vtheta \mid \D_t; \mW = \mW_*) = \mathtt{LA}(g_\vtheta, \mathcal{C}_t)\)  \Comment{Consisting of both the training (via SGD) and the Laplace approx.\ of the PEFT weights}
    \State \(p(g_t \mid \D_t) = \mathtt{LLA}(p(\vtheta \mid \D_t; \mW = \mW_*))\)  \Comment{Eq.\ \eqref{eq:lla_marginal}}
    \State \(\vx_t = \argmax_{\vx \in \Dtest} \alpha(p(g_t(c(\vx)) \mid \D_t))\)
    \State \(\D_{t+1} = \D_t \cup \{ (\vx_t, f(\vx_t)) \}\)
    \State \(\Dtest = \Dtest \setminus \{ \vx_t \}\)
    \EndFor
    \State \Return \(\argmax_{(\vx, f(\vx)) \in \D_{T+1}} f(\vx)\)
  \end{algorithmic}
\end{algorithm}

\subsection{Datasets}
\label{app:subsec:datasets}

Here, we expand the details of the datasets mentioned in the main text:
\begin{itemize}
  \vspace{-0.85em}
  \setlength\itemsep{0em}
  \item \textbf{Redoxmer \citep{agarwal2021}:} \num{1407} molecules.
  \item \textbf{Solvation \citep{agarwal2021}:} the same molecules as in Redoxmer.
  \item \textbf{Laser \citep{laser2023}:} \num{10000} molecules, subsampled without replacement from the original \num{182858} molecules.
  \item \textbf{Photovoltaics \citep{lopez2016cleanenergy}:} \num{10000} molecules, subsampled without replacement from the original \num{2320648} molecules.
  \item \textbf{Kinase \citep{graff2021}:} \num{10449} molecules.
  \item \textbf{Photoswitches \citep{griffiths2022photoswitches}:} \num{392} molecules.
  \item \textbf{Multi-Redox \citep{agarwal2021}:} \num{1407} molecules.
  \item \textbf{Multi-Laser \citep{laser2023}:} \num{10000} molecules, subsampled without replacement from the original \num{182858} molecules.
\end{itemize}

For all virtual libraries/datasets above, we have a ground-truth label for every molecule.
They are obtained (by the corresponding original authors) from physically founded simulations.
For instance, quantum chemistry or density functional theory for the materials datasets and molecular dynamics docking for drug discovery.
Refer to the respective paper for the details on how the ground-truth labels are computed.

\subsection{Training}
\label{app:subsec:trainings}

\subsubsection{Fixed-feature surrogates}

The following are the training details of the surrogates we used in \cref{sec:llm_feat}.
Note that for the LLM features, we average the last transformer embedding of shape \texttt{(batch\_size, seq\_len, embd\_dim)} over the second axis, while ignoring padding and \texttt{EOS} tokens.
We used HuggingFace's \texttt{transformers} library \citep{wolf2019huggingface} for all the LLM-related objects that appeared in this paper.

\paragraph*{GP} We use BoTorch \citep{balandat2020botorch} to construct the surrogate function.
The Tanimoto kernel is taken from Gauche \citep{gauche}.
To optimize the marginal likelihood, we use Adam \citep{kingma2014adam} with learning rate \num{0.01} for \num{500} epochs.

\paragraph*{LA} Our implementation is based on the \texttt{laplace-bayesopt} package \url{https://github.com/wiseodd/laplace-bayesopt}.
The neural net used is a \num{2}-hidden-layer multilayer perceptron with \(50\) hidden units on each layer along with the tanh activation function for fingerprint features (except for the Photoswitches dataset since we observe a vanishing gradient problem).
Otherwise, we use the ReLU activation function.
We optimize the network with Adam with learning rate \num{1e-3} and weight decay \num{5e-4} for \num{500} epochs with a batch size of \(20\).
We anneal the learning rate with the cosine annealing scheme \citep{loshchilov2017cosine}.
The Laplace approximation is done \emph{post hoc} and we tune the prior precision with the marginal likelihood for \(100\) iterations.
The Hessian is approximated with a Kronecker structure \citep{ritter2018laplace} except for the Llama-2-7b features---for the latter case, we found that the diagonal structure performs better.

\subsubsection{Finetuned surrogates}

The following are the training details of the surrogates we used in \cref{sec:lapeft}, i.e.\ the LA on LoRA weights \citep{hu2022lora,yang2023loralaplace}.
We jointly trained the LoRA and the regression head with AdamW with batch size \num{16}, learning rate \num{3e-4} and \num{1e-3} for the LoRA weights and the regression head's weights respectively (except for Photoswitch where we used \num{3e-3} and \num{1e-2} respectively), and weight decay \num{0.01} for \num{50} epochs.
Then, we further optimized only the regression head with the same hyperparameters for \num{100} epochs.

We used LoRA with rank \num{4} without bias on the key and value attention weights, following the original paper, except for GPT2 where we applied LoRA on all attention weights since the GPT2 implementation we used coupled all the attention weights into a single weight matrix.
Moreover, the \(\alpha\) (scaling) hyperparameter was set to \num{16}.
Additionally, we used dropout on LoRA with probability \num{0.1}.
See \citet{hu2022lora} for the explanation of these hyperparameters.
We used HuggingFace's \texttt{PEFT} library \citep{peft}.

The Laplace approximation was done on all LoRA's weights and the head's weights.
We used Kronecker-factored Hessian and optimized the layerwise prior precisions \citep{daxberger2021laplace} with \emph{post hoc} marginal likelihood for \num{200} iterations.

\subsection{Prompting}
\label{app:subsec:prompts}

We use the following prompts in our experiments (\cref{fig:prompts_assorted,fig:prompts_full}):
\begin{itemize}
  \vspace{-1em}
  \setlength\itemsep{0em}

  \item \textbf{just-smiles:} \texttt{``\{smiles\_str\}''}.
  \item \textbf{completion:} \texttt{``The estimated \{objective\_str\} of the molecule \{smiles\_str\} is: ''}.
  \item \textbf{single-number:} \texttt{``Answer with just numbers without any further explanation! What is the estimated \{objective\_str\} of the molecule with the SMILES string \{smiles\_str\}?''}.
  \item \textbf{naive:} \texttt{``Predict the \{objective\_str\} of the following SMILES: \{smiles\_str\}!''}.
        \vspace{-1em}
\end{itemize}

The variable \texttt{smiles\_str} equals the SMILES representation of the molecule at hand, e.g.\ ``OS(=O)(=O)O'' for sulfuric acid.
The variable \texttt{obj\_str} has a value the textual description of the problem at hand: ``redox potential'' for Redoxmer, ``solvation energy'' for Solvation, ``docking score'' for Kinase, ``fluorescence oscillator strength'' for Laser, ``Pi-Pi* transition wavelength'' for Photoswitches, ``power conversion efficiency'' for Photovoltaics, ``redox potential and solvation energy'' for MultiRedox, and ``fluorescence oscillator strength and electronic gap'' for MultiLaser.

The idea behind this selection of prompts is to cover all the expected inputs of LLMs.
For example, GPT-2 and Llama-2 are trained as next-word-prediction models.
So, the ``completion'' prompt might be the most suitable prompt template for them.

\subsection{In-context-learning baselines}
\label{app:subsec:incontext_baselines}

In \cref{subsec:icl_baseline}, we compared Bayesian surrogates with BO-LIFT, which uses in-context learning in conjunction with non-Bayesian uncertainty estimation \cite{ramos2023bayesopticl}.
To perform prompting, we use their top-$k$ completions template, which selects $k$ previously-seen examples with known properties and appends the SMILES string of the query molecule:

\texttt{Q: Given smiles \{smiles\_str\_1\}, what is \{obj\_str\}? A: \{\(f(\vx_1)\)\} \#\#\# ... \#\#\# Q: Given smiles \{smiles\_str\_k\}, what is \{obj\_str\}? A: \{\(f(\vx_k)\)\} \#\#\# Q: Given smiles \{smiles\_str\_query\}, what is \{obj\_str\}? A:}

The $k$-shot examples above are chosen based on the similarity and diversity of their Ada embeddings (part of OpenAI's GPT-4).
Then, $n$ completions to the prompt are sampled from the LLM, to compute the mean and variance of \(g_t(\vx)\).
We use Thompson sampling to make BO-LIFT comparable to the other methods.
The values of $k$ and $n$ were set to default values in the BO-LIFT code base, which are both \num{5}.

During our experiment with BO-LIFT (using the authors' code \url{https://github.com/ur-whitelab/BO-LIFT} to prompt \boliftgpt{}), we found that the monetary cost of performing it on the full Redoxmer dataset was already over USD 60 at the third BO iteration \(t = 3\).
Extrapolating, this means a BO run with \(T = 100\) for five different random seeds---as done for other methods considered in this work---will amount to roughly USD \num{20000}.
For this reason, in \cref{subsec:icl_baseline}, we only used a subsampled Redoxmer dataset \(\abs{\Dtest} = 200\) for only \(T = 15\) BO rounds (repeated over five random seeds).
Even in this setup, the cost of BO-LIFT was already USD 75.81.

To circumvent this monetary cost, we also tried ICL using the ``chat'' version of \boliftllama{}, the most capable free model we could run on a single GPU.
However, we also found it difficult to run this model on our entire Redoxamer dataset because it was projected to take approximately \num{100}
hours for \(T = 200\) BO rounds.
This is another reason why we decided to do the experiment in \cref{subsec:icl_baseline} with only the subsampled Redoxmer dataset.

All in all, this shows that contrary to popular belief, using principled Bayesian surrogates in a BO over molecules is actually cheaper than using ICL-based non-Bayesian surrogates.
As we have shown in the main text, this is because one does not need very large, state-of-the-art LLMs to do BO.
One only needs to use small, domain-specific LLMs like that of \citet{christofidellis2023unifying}.


\section{Additional Results}
\label{app:sec:results}

\begin{figure}
	\begin{center}
		\includegraphics[width=0.475\textwidth]{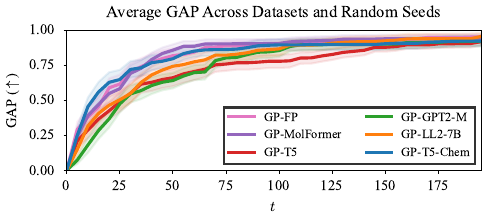}
		\hfill
		\includegraphics[width=0.475\textwidth]{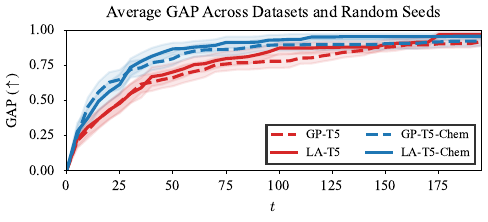}
	\end{center}

	\vspace{-1.5em}

	\caption{
		\textbf{Left:} Average performance of GP with the Tanimoto kernel on transformer features.
		\textbf{Right:} GP versus LA surrogates in terms.
	}
	\label{fig:gap_gp}
\end{figure}

\begin{figure*}
	\centering
	\includegraphics[width=\linewidth]{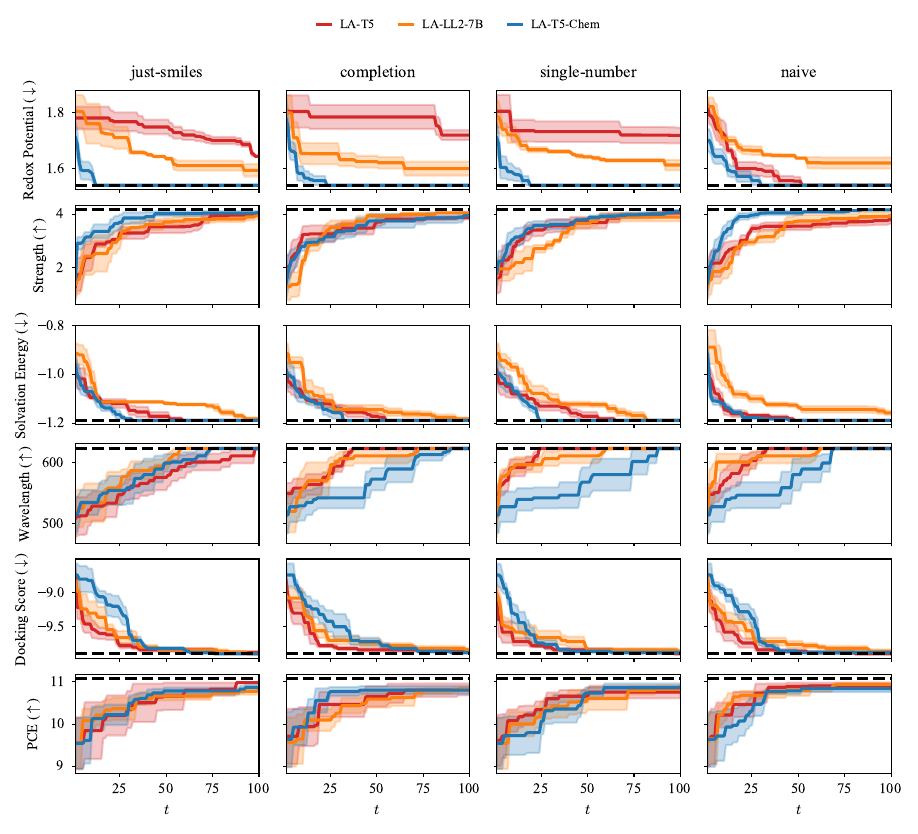}

	\vspace{-1em}
	\caption{
		Different prompts.
	}
	\label{fig:prompts_full}
\end{figure*}

In this section, we present additional and/or detailed results to supplement the results in the main text.

\subsection{Laplace vs.\ GP}
\label{app:subsec:la_vs_gp}

We show the performance of the GP (with the Tanimoto kernel) surrogate vis-\`{a}-vis LA in \cref{fig:gap_gp}.
Our conclusion is two-fold:
(i) We observe a similar ordering as in the main text---chemistry-specific LLMs/transformers are best, although here the performance difference is less pronounced.
(ii) The LA surrogate is generally better than the GP one in our tasks.

\subsection{Acquisition functions}
\label{app:subsec:acqfs}

\begin{figure*}
	\centering
	\subfloat[Fixed]{\includegraphics[width=0.475\linewidth]{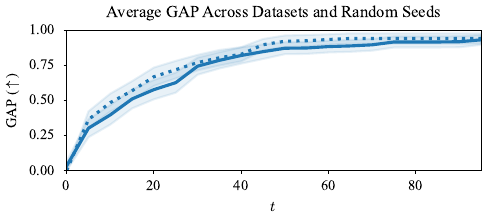}}
	\subfloat[Finetuned]{\includegraphics[width=0.475\linewidth]{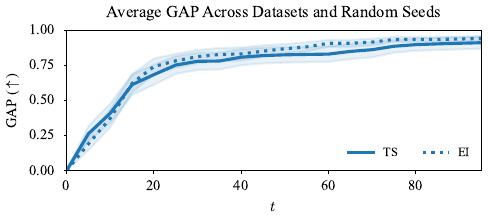}}

	\caption{
		Summarized BO performance with Thompson sampling and expected improvement.
	}
	\label{fig:acqf_gap}
\end{figure*}

\begin{figure*}
	\centering
	\includegraphics[width=\linewidth]{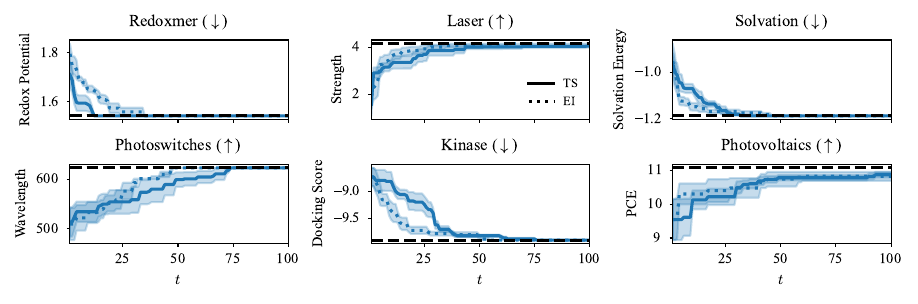}

	\vspace{-1em}
	\caption{
		Fixed-feature BO (\latfivechem{}) with different acquisition functions---Thompson sampling (TS) and expected improvement (EI).
	}
	\label{fig:acqf_fixed}
\end{figure*}

\begin{figure*}
	\centering
	\includegraphics[width=\linewidth]{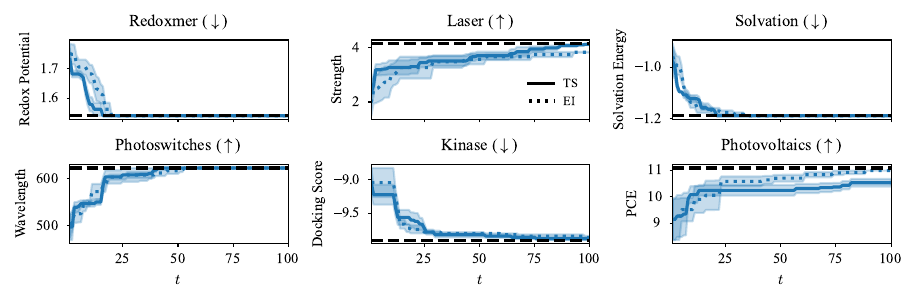}

	\vspace{-1em}
	\caption{
		Finetuning \latfivechem{} with different acquisition functions---Thompson sampling (TS) and expected improvement (EI).
	}
	\label{fig:acqf_ft}
\end{figure*}

In \cref{fig:acqf_fixed,fig:acqf_ft,fig:acqf_gap} we show the comparison between expected improvement (EI) and Thompson sampling (TS) acquisition functions.
We found that, on aggregate, the difference between EI and TS is insignificant (\cref{fig:acqf_gap}, see the error bars).
Meanwhile, in some individual cases, their difference can be significant in both directions.
So, our suggestion is to do a grid search on a proxy problem whenever possible; otherwise, simply use whichever acquisition function is available.

\subsection{Prompts}
\label{app:subsec:prompts_exps}

In \cref{fig:prompts_full}, we show the effect of the choice of prompt on the BO performance in the fixed-feature (\cref{fig:one}(a)) case.
We found that prompts do affect BO performance in all datasets considered.
Specifically for \latfivechem{}, the just-smiles prompt template is best.
This is consistent with how \latfivechem{} was finetuned, i.e., using just the SMILES strings as the inputs \citep{christofidellis2023unifying}.
Surprisingly, \lallama{} does not gain substantial benefits when different prompt templates are used, even when the ``completion'' prompt template is used.
This indicates that while \lallama{} can convincingly answer natural language questions, it does not contain useful information for chemistry (see \cref{fig:one}).
Finally, we did not see a clear improvement pattern in BO with different prompts using \latfive{}.
All in all, these findings further support our conclusion that one should use domain-specific LLMs (e.g.\ \latfivechem{}) for doing BO.

\begin{figure}
	\centering
	\includegraphics[width=\linewidth]{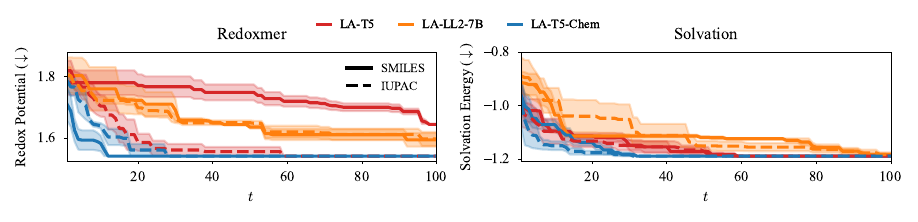}

	\vspace{-1em}
	\caption{
		SMILES vs.\ IUPAC.
	}
	\label{fig:smiles_vs_iupac}
\end{figure}

\begin{figure*}
	\centering
	\includegraphics[width=\linewidth]{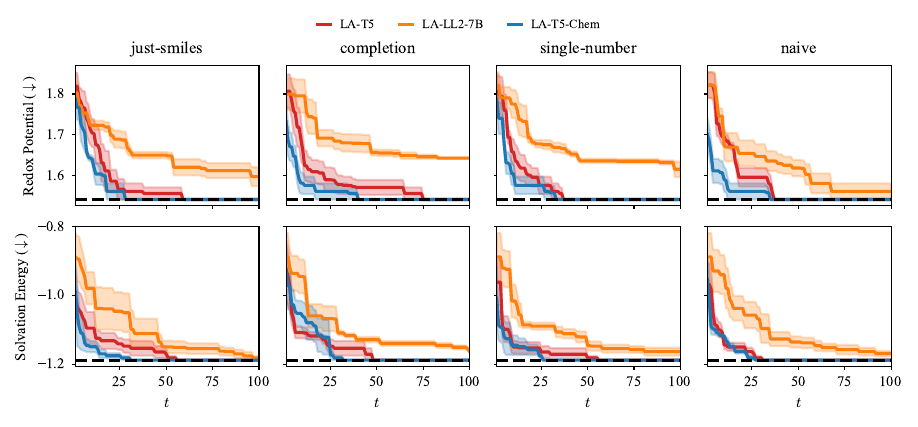}

	\vspace{-1em}
	\caption{
		IUPAC with different prompts.
	}
	\label{fig:iupac_prompts}
\end{figure*}

\subsection{Textual representations}
\label{app:subsec:text_repr}

As for which textual representation of the molecule to use, we compare SMILES with IUPAC representations in \cref{fig:smiles_vs_iupac}.
We found that \lallama{} is insensitive to this.
Meanwhile, observed significant effects on both \latfive{} and \latfivechem{}.
Thus, we can conclude that it might be worth trying different molecular string representations (IUPAC, SMILES, SELFIES, etc.) when using \latfivechem{}.
Moreover, in \cref{fig:iupac_prompts}, we show the impact of prompting when IUPAC representations are used.
Our findings are similar to the ones with SMILES.
Especially \lallama{} is largely insensitive to prompting.

\subsection{Computational costs}
\label{app:subsec:costs}

\begin{figure*}
	\centering
	\includegraphics[width=\linewidth]{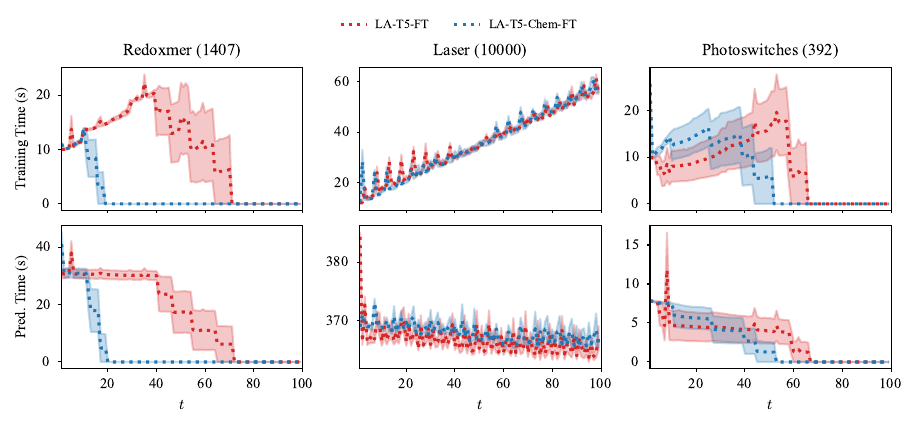}

	\vspace{-1em}
	\caption{
		Wall clock time (in seconds) per BO iteration of the finetuned surrogates.
		Numbers in parentheses are numbers of test points \(\vert \Dtest \vert\).
		Notice that the costs are roughly linear in the number of test points, indicating that forward passes over \(\Dtest\) take the bulk of the computation.
		\latfivechem{} has a lower wall-clock time on average since it tends to terminate faster than \latfive{}.
		E.g., some of the BO runs across random seeds have been done and thus have zero wall-clock time.
	}
	\label{fig:timing}
\end{figure*}

We present the computational cost of performing finetuning in \cref{fig:timing} on datasets of various magnitudes (\num{392}, \num{1407}, \num{10000}).
We use a single NVIDIA A40 and NVIDIA RTX6000 GPUs for the laser and the rest of the finetuning experiments, respectively.\footnote{For fixed-feature computation, we run the BO loop on consumer-level laptop CPU.}
Since the number of training points is constant across those datasets (equals to \(t\) plus \(10\), the latter is the size of the initial training set \(\D_1\)), the time needed for each BO iteration is almost exclusively influenced by the prediction phase of the BO iteration (\cref{alg:bo_molecules}, line 3).
Indeed, we see an almost exact scaling in terms of the size of \(\Dtest\).
This is encouraging since finetuning/training is \emph{not} the bottleneck, contrary to popular belief.

\end{appendices}

\end{document}